\documentclass[conference]{IEEEtran}
\IEEEoverridecommandlockouts                              

\usepackage{times}
\usepackage[numbers]{natbib}
\usepackage{multicol}
\usepackage[bookmarks=true]{hyperref}
\usepackage{cite}
\usepackage{amsmath,amssymb,amsfonts}
\usepackage{algorithmic}
\usepackage{graphicx}
\usepackage{textcomp}
\usepackage{xcolor}
\usepackage{booktabs}
\usepackage{todonotes}
\usepackage{amsmath}
\usepackage{amssymb}
\usepackage{bm}

\makeatletter
\newcommand{\newparallel}{{\mathrel{\mathpalette\new@parallel\relax}}}
\newcommand{\new@parallel}[2]{%
  \begingroup
  \sbox\z@{$#1T$}
  \resizebox{!}{\ht\z@}{\raisebox{\depth}{$\m@th#1/\mkern-5mu/$}}%
  \endgroup
}
\makeatother

\renewcommand{\arraystretch}{1.5}  

\title{\LARGE \bf Flying Hand: End-Effector-Centric Framework for Versatile Aerial Manipulation Teleoperation and Policy Learning}

\author{Guanqi He*$^{\dagger}$, Xiaofeng Guo*$^{\dagger}$, Luyi Tang$^{\dagger}$, Yuanhang Zhang$^{\dagger}$, Mohammadreza Mousaei$^{\dagger}$, Jiahe Xu$^{\dagger}$, \\
Junyi Geng$^{\ddagger}$, Sebastian Scherer$^{\dagger}$, Guanya Shi$^{\dagger}$%
\thanks{$^{*}$Equal Contribution, Alphabetical Order. $^{\dagger}$Robotics Institute, Carnegie Mellon University, Pittsburgh PA 15213, USA. $^{\ddagger}$Department of Aerospace Engineering, Pennsylvania State University, University Park, PA, 16802, USA.}
}

\begin{document}

\maketitle
\thispagestyle{empty}
\pagestyle{empty}

\begin{abstract}

Aerial manipulation has recently attracted increasing interest from both industry and academia. Previous approaches have demonstrated success in various specific tasks. However, their hardware design and control frameworks are often tightly coupled with task specifications, limiting the development of cross-task and cross-platform algorithms. Inspired by the success of robot learning in tabletop manipulation, we propose a unified aerial manipulation framework with an end-effector-centric interface that decouples high-level platform-agnostic decision-making from task-agnostic low-level control. Our framework consists of a fully-actuated hexarotor with a 4-DoF robotic arm, an end-effector-centric whole-body model predictive controller, and a high-level policy. The high-precision end-effector controller enables efficient and intuitive aerial teleoperation for versatile tasks and facilitates the development of imitation learning policies. Real-world experiments show that the proposed framework significantly improves end-effector tracking accuracy, and can handle multiple aerial teleoperation and imitation learning tasks, including writing, peg-in-hole, pick and place, changing light bulbs, etc. We believe the proposed framework provides one way to standardize and unify aerial manipulation into the general manipulation community and to advance the field. Project website: \url{https://lecar-lab.github.io/flying_hand/}.

\end{abstract}


\IEEEpeerreviewmaketitle

\section{Introduction}
\label{sec: Intro}

Uncrewed Aerial Manipulators (UAMs), which target complex tasks at high altitudes \citep{suarez2020benchmarks}, hold significant potential to reduce human labor in many elevated operations, such as changing light bulbs on tall towers, inspecting aircraft wings or turbine blades, and maintaining or painting bridges, which are not only costly but also pose substantial risks to human safety. Previous works have demonstrated the ability to achieve different specific complex aerial manipulation tasks, including drawing calligraphy~\citep{guo2024flying}, grasping \citep{mellinger2011design}, perching \citep{ji2022real}, drilling \citep{ding2021design}, etc. However, most previous works have been tailored to specific tasks, developing unique platforms and algorithms accordingly, lacking the ability to handle different types of tasks. In real-world scenarios, manipulation tasks can be complex and typically consist of multiple sub-tasks. For example, changing a light bulb can involve several motion primitives, including interaction, grasping, insertion, and rotation. This raises a requirement for a more general-purpose versatile aerial manipulation system, which essentially requires a \textbf{versatile aerial manipulation framework} to handle multiple tasks. 

\begin{figure}[!t]
    \centering
    \includegraphics[width=0.99\linewidth]{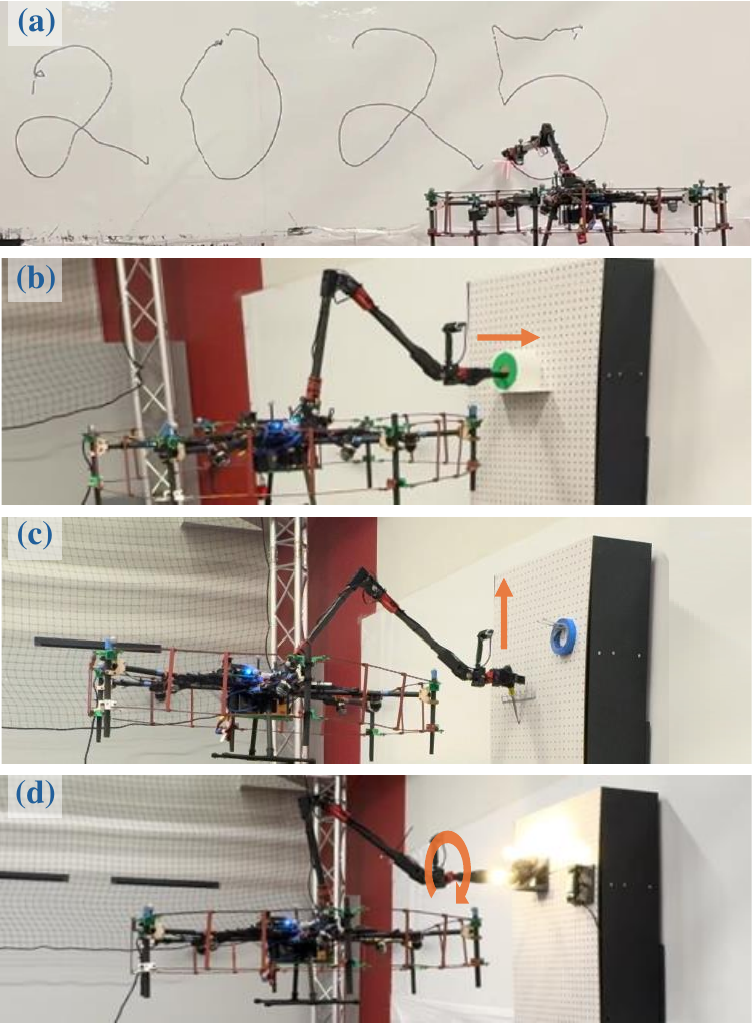}
    \caption{The proposed framework and system can accomplish multiple typical aerial manipulation tasks precisely and robustly, such as (a) writing "2025", (b) peg-in-hole, (c) pick-and-place, and (d) changing light bulbs.}
    \label{fig: concept}
\end{figure}

In robotic manipulation \citep{mason2018toward}, the end-effector-centric (ee-centric) approach is widely used. It defines tasks and policies \citep{khatib1987unified, kroemer2021review} in Cartesian space instead of specific robotic arm configuration space. By effectively decoupling high-level policies from low-level control, it enables embodiment-agnostic policies \citep{team2024octo}, \citep{chi2024umi}, \citep{song2020grasping} and policy-agnostic low-level controllers development \citep{lewis2003robot}, enhancing framework versatility, cross-embodiment adaptability, and algorithm reuse capability \citep{yang2024pushing}. Although shown the advantage of versatility in the manipulation field, applying the ee-centric paradigm to aerial manipulation systems presents significant challenges due to the UAM's floating-base dynamics and the coupling effects between the UAV and the manipulator.

\begin{figure*}[h!]
    \centering
    \includegraphics[width=\linewidth]{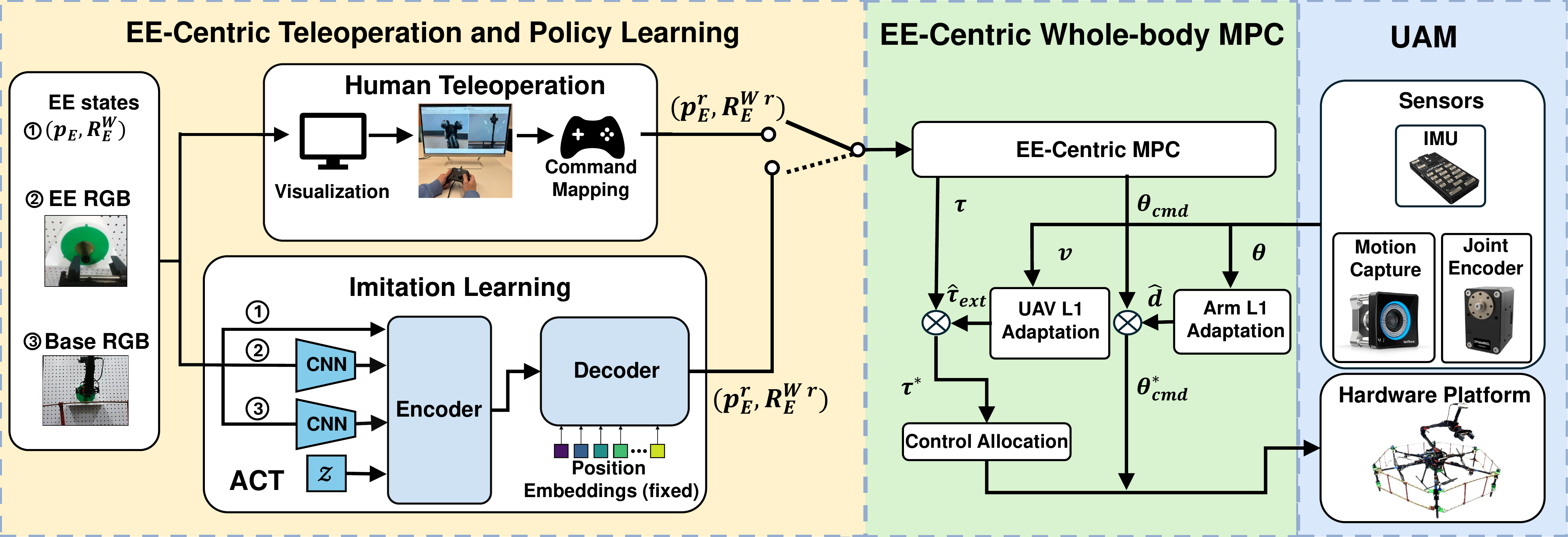}
    \caption{The proposed end-effector-centric aerial manipulation framework includes the UAM platform, the ee-centric whole-body MPC, and the high-level policy including an ee-centric teleoperation interface, and an imitation learning-based framework using Action Chunk with Transformer (ACT) \citep{zhao2023learning}. The high-level policy, either the human teleoperation or learned autonomous policy sends the target end-effector state to ee-centric MPC which then generates motor commands for the UAM platform to execute.}
    \label{fig: pipeline}
\end{figure*}

In this work, inspired by the success of ee-centric interfaces in classic manipulation, we reformulate aerial manipulation from the \textbf{manipulation} perspective by proposing a versatile aerial manipulation framework with the ee-centric interface to address various aerial manipulation tasks.
The framework consists of a versatile aerial manipulation platform capable of executing multiple tasks, a policy-agnostic controller that precisely tracks the target end effector state, and an ee-centric policy module responsible for generating target ee states. Specifically, we develop a fully-actuated hexarotor with a 4 DoF robotic arm, providing a sufficiently large workspace and wrench space for diverse tasks. We then develop an ee-centric whole-body Model Predictive Controller (\textbf{ee-centric MPC}) that precisely tracks the target end-effector state, even in the presence of model uncertainties. Moreover, to bring human cognitive skills into policy development and benefit from the ee-centric interface, we develop an ee-centric teleoperation interface and an imitation-learning-based framework to acquire autonomous policy learned from human demonstration. To the best of our knowledge, this is the first imitation learning-based framework for aerial manipulation. Real-world experiments show that the proposed framework achieves high-precision end-effector tracking and enables a wide range of aerial manipulation tasks, including aerial writing, peg-in-hole, pick and place, light bulb replacement, etc., as shown in Fig. \ref{fig: concept}. It also demonstrates how this modular and standardized ee-centric framework effectively decouples the high-level policy from the low-level controller, which enables seamless integration of existing standard high-level policy modules from the broader manipulation community, such as teleoperation and imitation learning, into the field of aerial manipulation. We believe the proposed framework provides a step toward standardizing and unifying aerial manipulation into the broader manipulation community, advancing the field toward greater versatility and generalization.

In summary, our contributions are:

1). We reformulated the aerial manipulation problem within the unified manipulation framework, consisting of a UAM system, a controller encapsulated by the ee-centric interface, and a high-level policy.

2). We developed an end-effector-centric whole-body MPC for aerial manipulation that precisely tracks the targeted end-effector state while maintaining robustness against disturbances through L1 adaptation.

3). We developed an ee-centric teleoperation system and an imitation-learning-based autonomous system that learns from human teleoperation demonstration.

4). Rich real-world experiments demonstrated the versatility of the proposed framework, the effectiveness of the user-friendly teleoperation interface, and the potential to incorporate learning-based policies and other manipulation policies.

\section{Related Works}
\label{sec: related_works}

\subsection{Aerial Manipulation}
\label{sec: related_works_am}

There have been many research efforts exploring aerial manipulation for various kinds of tasks \citep{ollero2021past}. Based on the motion primitives they require, common aerial manipulation tasks can be categorized into: 1). Aerial Interaction, which requires maintaining contact with external objects, for tasks such as inspection \citep{bodie2019omnidirectional} \citep{bodie2020active} \citep{guo2024aerial}, aerial writing \citep{lanegger2022aerial} \citep{tzoumanikas2020aerial}, \citep{10342145} or pushing a target \citep{brunner2022energy}. Researchers mostly developed a point-contact arm such as a rigid rod, and proposed the hybrid motion-force control framework, although achieving high-precision tracking performance, struggles to handle scenarios requiring grasping; 2) Aerial grasping \citep{meng2022aerial}, where previous works mainly focused on designing different custom end-effectors, such as claw \citep{roderick2021bird} or soft gripper \citep{fishman2021dynamic}. Some work also showed amazing results achieving high-speed grasping, or grasping moving objects \citep{ubellacker2024softgrasp}, but sacrificed payload capacity or precision due to specialized hardware designs. 3). Aerial insertion. Typical work includes \citep{schuster2019screwing} where they propose a specific hole searching policy for bolt screwing tasks, and \citep{wang2023millimeter} where they achieved mm-level peg-in-hole task; 4). Manipulate articulated objects such as doors \citep{su2023sequential}, or valves \citep{brunner2022planning}. In general, although different works have shown success on different specific tasks, the specific system design and algorithm development make the same hardware and algorithm hard to deploy to different tasks, reducing its potential for practical long-horizon versatile aerial manipulation tasks. In our work, we target all these four types of aerial manipulation tasks, developing a versatile framework to handle all of them.

\subsection{Manipulation with EE-Centric Interface}

Defining tasks or commands using ee-centric approaches is widely adopted in general manipulation fields, as it is more intuitive and can be cross-embodiment. For example, \citep{baberg2016ee-centric} developed a teleoperation interface to enable full control of the end-effector pose. The Universal Manipulation Interface (UMI) \citep{chi2024umi} \citep{ha2024umilegs} demonstrates a data-collection and policy-learning framework that allows direct skill transfer from in-the-wild human demonstrations to multiple robot embodiments. Their system employs a hand-held gripper and carefully designed hardware-agnostic policies, showcasing the potential for ee-centric solutions in multi-platform scenarios. Similarly, other mobile manipulation strategies, such as N2M2 \citep{honerkamp2022n2m2} \citep{honerkamp2021n2m2} and HarmonicMM \citep{yang2024harmonicmm}, reduce the operator’s burden by extracting feasible base motions from end-effector trajectories. But they generally remain limited to ground robots. Although several aerial manipulation studies have adopted end-effector-centric methods \citep{guo2024flying} \citep{bodie2021dynamic}, they have primarily focused on developing controllers to track specified end-effector trajectories without a systematic framework that tackles various tasks comprehensively. In our work, we propose a unified framework with the ee-centric interface for versatile aerial manipulation tasks.

\subsection{Teleportation and Imitation Learning}

Developing a robust and practical autonomous aerial manipulation policy is extremely challenging due to complex real-world environments and high precision and safety requirements. Moreover, policies are typically designed to handle specific tasks and lack the generality to handle unexpected conditions. Therefore, teleoperation, which takes human effort into the loop for policy development, attracts researchers' interest as a practical solution. For example, \citep{allenspach2022towards} developed the UAM with a fully actuated UAV with 0 DoF arm and controlled the end effector directly by teleoperation, but their method is highly coupled with the specific UAM design and the system suffers for versatile tasks due to the workspace limitation. In most structured UAM teleoperation works, users control each DoF separately \citep{yashin2019aerovr} or explicitly switch to different modes during different phases \citep{coelho2021whole}. Both of these increase the human teleoperator's burden and require the teleoperator to have a rich understanding of the specific hardware system. Moreover, even if incorporated with human input, the previous works still have not shown vast versatility in different tasks and scenarios.

Recently, imitation learning (IL) has demonstrated significant potential for autonomous policy learning due to its high data efficiency, straightforward framework, and outstanding performance. Recent progress in both systems, such as ALOHA \citep{zhao2023learning} and mobile ALOHA \citep{fu2024aloha}, and algorithms, such as ACT \citep{zhao2023learning}, and diffusion policy \citep{chi2023diffusion}, facilitate the success of long-horizon, contact-rich, complex manipulation tasks. However, there is no precedent to incorporate such IL-based policy into aerial manipulation fields due to the lack of a mature demonstration collection system such as a well-development teleoperation system, and the lack of a proven framework. In this work, we develop an intuitive teleoperation system using the ee-interface framework. It also helps to collect human demonstration data, enabling us to develop an imitation learning-based policy for autonomous aerial manipulation.

\section{System Overview}
\label{sec: system_overview}

Our aerial manipulation system is designed to enable precise and versatile operations. The system incorporates an end-effector-centric (ee-centric) interface to decouple high-level decision-making from low-level control, increasing the framework's versatility. As shown in Fig. \ref{fig: pipeline}, our system consists of an aerial manipulator platform, an ee-centric whole-body MPC, and an ee-centric policy module. The hardware platform consists of a fully-actuated hexarotor and a 4 DoF robotic arm. The platform has a large enough workspace and wrench space for different tasks. A motion capture system and onboard IMUs are used for drone state estimation. The joint encoders are used to get arm joint angles and the end-effector states are then calculated based on forward kinematics. The ee-centric whole-body MPC reads the ee target from high-level policy and generates the reference trajectory and reference control for both the UAV and the robotic arm. An L1 online adaptation control term is designed to further improve the tracking performance. The UAV control commands are then sent to the control allocation to generate the motor commands for the UAV to execute. At the highest level, the ee-centric policy module gets current observations and generates the target ee states online without the need to consider the specific platform jointly. We developed two high-level policy modules. The first one is the ee-centric teleoperation interface that allows human users to directly control the end-effector pose. Based on the human demonstration, we adopted an imitation-learning-based method: Action Chunk with Transformer(ACT) \citep{zhao2023learningfinegrainedbimanualmanipulation}, to learn an autonomous policy. The following sections will introduce the developed modules accordingly.

\section{Hardware Design}
\label{sec: hardware}

\subsection{Fully-Actuated UAV}
The foundation of the system is a fully-actuated hexarotor UAV capable of independently generating six-dimensional forces and torques. This capability allows precise control of position and orientation, which is essential for executing complex aerial manipulation tasks. The robust design ensures stability in dynamic environments, while ensuring high-precision end-effector tracking. We used Tarot680 as the drone base, 6 KDE 4215XF motor with a 12-inch 2-blade propeller as our driving force, LiPo batteries for on-board power supply, an on-board computer Intel Nuc for on-board computation, and a customized PX4 autopilot for low-level flight control and information processing.

\subsection{Manipulator}
The UAV integrates a 4-DOF robotic manipulator optimized for versatile and precise task execution. The arm features three pitch joints and one roll joint, driven by Dynamixel XM540 and XM430 servo. Its configuration allows high-precision operations. The system achieves whole-body manipulation capabilities by combining the UAV’s actuation with the manipulator, enhancing task execution across diverse scenarios. The manipulator includes a modular end-effector, allowing interchangeable tools for specific tasks. For instance, a two-finger gripper with replaceable tips enables precision handling, while a circular gripper is ideal for changing light bulbs. 

\subsection{Perception}
In this work, we mainly focus on the in-door environment. We use both the motion capture system and the PX4 onboard IMU for drone state estimation. 
The motion capture system provides the UAV's position and can be replaced with other localization methods, such as SLAM.
The manipulator arm joint angles are estimated by the joint encoders and the end-effector states are online calculated based on forward kinematics (FK). 

To further improve drone perception for teleoperation and autonomous policy development, we equip the aerial manipulator with two RealSense RGBD cameras. One camera is mounted on the UAV base to capture a broad view of the entire workspace, while the other is positioned near the end-effector to deliver detailed close-up views of the target area. This dual-camera setup ensures teleoperators and vision-based policies can maintain precise control and situational awareness during complex manipulation tasks.

\begin{figure}
    \centering
    \includegraphics[width=0.99\linewidth]{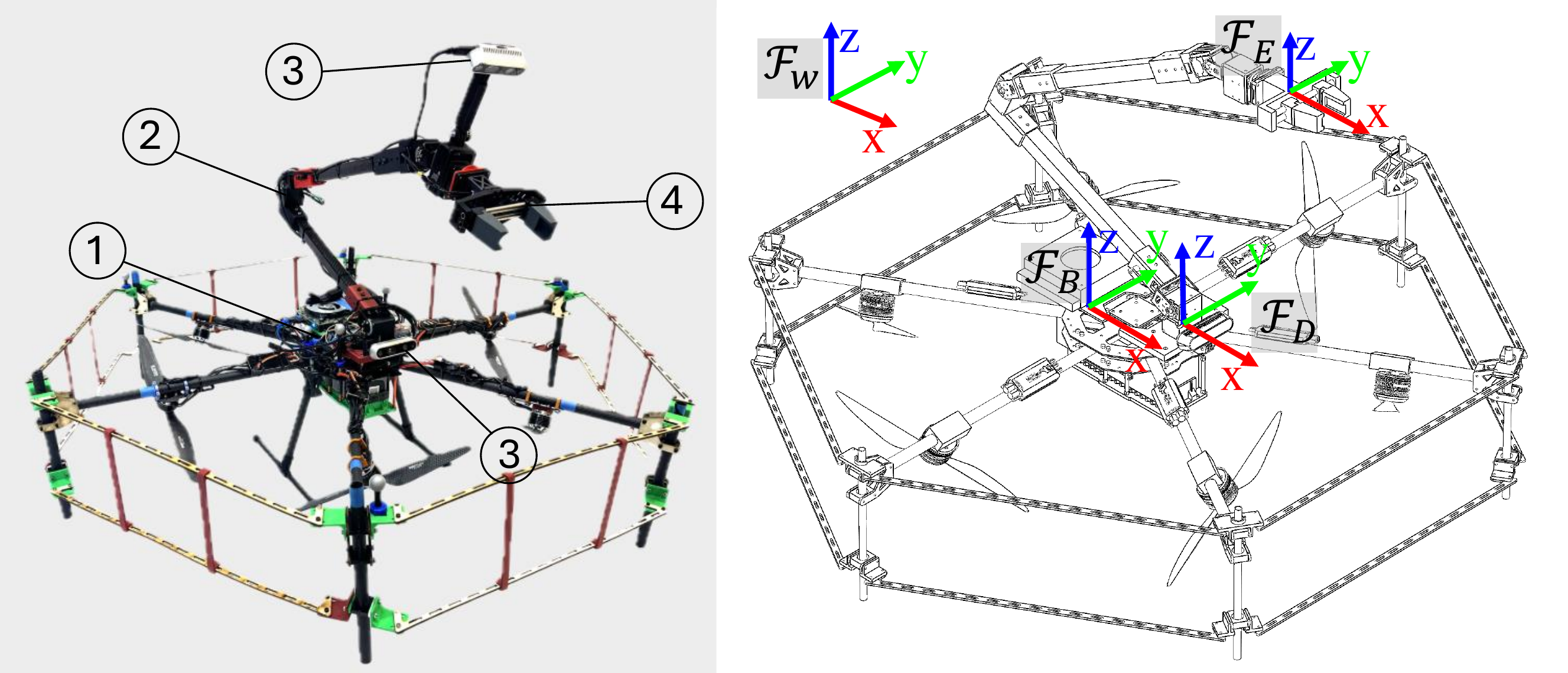}
    \caption{UAM hardware system design, illustrating the key components: (1) fully-actuated hexarotor as the base structure, (2) 4 Dof manipulator, (3) Intel RealSense cameras for vision-based perception and feedback, and (4) end-effector gripper for object interaction. The frame notations in the right diagram represent the coordinate axes associated with the system.}
    \label{fig:uam_design}
\end{figure}

\section{System Modeling}
\subsection{Frames and Notation}
The frames depicted in the Fig \ref{fig:uam_design} are defined as follows: $\mathcal{F}_W$ is an inertial world frame with its $z$-axis opposite to the gravity vector, ensuring $\hat{z}_W$ points upward. $\mathcal{F}_B$ is rigidly attached to the UAV's body at its center of gravity, with axes $(x_B, y_B, z_B)$ aligned with the UAV body frame. $\mathcal{F}_D$ is the manipulator base frame, centered at the attachment point of the manipulator on the UAV. The transformation from $\mathcal{F}_B$ to $\mathcal{F}_D$ is defined by a constant translation $\mathbf{p}_D \in \mathbb{R}^3$ and a fixed orientation $\bm R_B^D \in SO(3)$. $\mathcal{F}_E$ is the end-effector frame with axes $(x_E, y_E, z_E)$, where $x_E$ is aligned with the roll axis of the $4$th joint of the manipulator, while $y_E$ remains horizontal.
Other symbols in the paper are
listed in Table \ref{tab:notation} for the convenience of the following discussion.

\begin{table}[h!]
\renewcommand{\arraystretch}{1.25}
\centering
\caption{Notation Overview}
\label{tab:notation}
\begin{tabular}{c c p{5cm}}
\toprule
\textbf{Symbol} & \textbf{Dimension} & \textbf{Description} \\
\midrule
$m$ & $\mathbb{R}$ 
  & Mass \\
$\bm J$ & $\mathbb{R}^{3 \times 3}$ 
  & Inertia tensor \\
$\bm p$, $\bm R_{B}^{W}$ & $\mathbb{R}^3$, $SO(3)$ 
  & UAV position expressed in $\mathcal{F}_W$ and orientation between $\mathcal{F}_B$ and $\mathcal{F}_W$ \\
$\bm v$ & $\mathbb{R}^3$ 
  & Generalized velocity, expressed in $\mathcal{F}_W$ \\
$\bm{\tau}$ & $\mathbb{R}^6$ &  Control wrench  \\
$\bm{\tau}_{\mathrm{ext}}$ & $\mathbb{R}^6$ & External disturbance wrench \\
$\bm{g}_W$ & $\mathbb{R}^6$ & Gravity vector in $\mathcal{F}_W$ \\
\midrule
$\bm p_E$, $\bm R_{E}^{W}$ & $\mathbb{R}^3$, $SO(3)$ 
  & End-effector position expressed in $\mathcal{F}_W$ and orientation between $\mathcal{F}_E$ and $\mathcal{F}_W$ \\
$\bm{\theta}$ & $\mathbb{R}^4$ & Current joint angles \\
$\bm{\theta}_{\mathrm{cmd}}$ & $\mathbb{R}^4$ & Commanded joint angles \\
$\bm d $ & $\mathbb{R}^4$ & Joint servo disturbance \\
$\bm{\zeta}$ & $\mathbb{R}^{3\times4}$ & DH parameter with the $i^{\mathrm{th}}$ joint component $[\theta_i, l_i, a_i, \alpha_i]^T$ \\
$\bm\beta$ & $\mathbb{R}$ & Joint motor delay constant \\
\bottomrule
\end{tabular}
\end{table}

\subsection{Fully-Actuated UAV Dynamics}


A fully-actuated UAV is adopted as the base of the aerial manipulator, which can generate six-dimensional force and torque independently. Let the generalized position of the UAV be represented as $ \bm q = [\bm p, \bm R^{W}_B] $, where $\bm p \in \mathbb{R}^3 $ represents the position of the UAV in the global coordinate frame, and $ \bm R^{W}_B \in SO(3) $ represents its orientation. The generalized velocity is denoted as $ \bm v = [\dot{\bm p}, \bm \omega] $, where $ \bm \omega \in \mathbb{R}^3 $ is the angular velocity.

The UAV dynamics can be formulated using Newton-Euler equations for rigid body motion as follows:

\begin{align}\label{uav_dyn}
\bm{M} \dot{{\bm{v}}} + \bm{C} {\bm{v}} + \bm{g} = {\boldsymbol{\tau}} + {\boldsymbol{\tau}}_{\mathrm{cxt}}
\end{align}
with inertia matrix $\bm{M}\in\mathbb{R}^{6\times 6}$, centrifugal and Coriolis term $\bm{C}\in\mathbb{R}^{6\times6}$, gravity wrench $\bm{g}\in\mathbb{R}^{6}$, control wrench $\bm{\tau} \in\mathbb{R}^{6}$ from the UAM actuators, and unknown external wrench $\bm{\tau}_{ext} \in\mathbb{R}^{6}$ from model mismatch and manipulator interaction.

Specifically, we have
\begin{align} 
    \bm{M} &= \mathrm{diag} \left( \begin{bmatrix} m \bm{I}_{3\times 3} & \bm{J} \end{bmatrix} \right), \\
    \bm{C} &= \mathrm{diag} \left( \begin{bmatrix} m [\bm{\omega}]_{\times} & -\bm{J} [\bm{\omega}]_{\times} \end{bmatrix} \right), \\
    \bm{g} &= m \ \mathrm{diag} \left( \begin{bmatrix} \bm R_{B}^{W} & \bm{0}_{3 \times 3} \end{bmatrix} \right) \bm{g}_W.
\end{align}
where $m$ is the vehicle mass, $\bm{J}$ is the moment of inertia at the vehicle center of mass in the body frame, $\bm{g}_W = [0, 0, g, 0, 0, 0]^\top$ is the gravitational acceleration in $\mathcal{F}_W$, and $[\bm{*}]_{\times}$ is the skew-symmetric matrix associated with vector $\bm{*}$. 

\subsection{Manipulator Kinematics}

In this work, the manipulator employs servos as joint actuators, which cannot accurately control joint torques directly. Therefore, only the kinematics of the manipulator is considered in the system modeling. The interaction between the manipulator and the fully-actuated UAV is treated as a disturbance and is compensated in real-time using L1 adaptive control.

We use the standard Denavit–Hartenberg (DH) convention \citep{denavit1955kinematic} to model the forward kinematics of our 4-DoF robotic arm. 
Under the DH formulation, the adjacent frame transformation $\bm T_i^{i-1}$ is characterized by four parameters $\theta_i$, $d_i$, $a_i$, $\alpha_i$, where the first one is the joint angle and the last three are pre-identified and fixed during robot movement.
Define DH parameter $\zeta_i = [\theta_i, l_i, a_i, \alpha_i]^\top \in \mathbb{R}^4$. 
The frame transformation from end-effector frame to arm base body frame can be written as 
\begin{equation}
 \bm{T}_E^D(\bm{\theta}; \bm{\zeta}) = \prod_{i=1}^{4} {\bm{T}^{i-1}_{i}(\theta_i; \zeta_i)}
\end{equation}

Then the transformation from world frame to end-effector frame can be computed as follows:

\begin{align}\label{arm_kin}
    \bm{T}_E^W &= \begin{bmatrix}
    \bm{R}^{W}_E &  \bm{p}^{W}_E \\
     \bm{0}_{1 \times 3} & 1
\end{bmatrix}  =  \bm{T}_B^W \cdot \bm{T}_D^B \cdot \bm{T}_E^D
\end{align} where $\bm{T}_B^W$ can be obtained from UAV odometery and $\bm{T}_D^B$ is a fixed transformation between the manipulator base and the UAV.

The accurate DH parameters are obtained through system identification. We collected motion data of the manipulator using a motion capture system and compute the DH parameters via least squares regression. The detailed parameter values are presented in Table~\ref{tab:sysid}. 

\subsection{Manipulator Motor Delay}
The servo motor dynamics are approximated as first-order systems to account for command-to-state delay. For the 4-DoF manipulator, the relationship between commanded joint angles $\bm{\theta}_{\mathrm{cmd}} \in \mathbb{R}^4$ and actual joint angles $\bm{\theta} \in \mathbb{R}^4$ is governed by:
\begin{equation}\label{arm_dyn}
    \mathrm{diag}({\bm\beta}) \dot{{\bm \theta}} + {\bm \theta} = {\bm \theta}_{\mathrm{cmd}} + \bm d
\end{equation}
where $\bm \beta \in \mathbb{R}$ are the joint-specific time delay constants, and $\bm d \in\mathbb{R}^4$ is the unknown disturbance in servo control. This formulation captures the transient response characteristics of each actuator. 
The motor delay coefficients ${\bm \beta}$ are identified alongside the DH parameters using least squares regression, with the results presented in Table~\ref{tab:sysid}.

\section{End-Effector-Centric Whole-Body Control with Online Adaptation}
\label{sec: mpc}

\label{sec: controller}

Given the over-actuated nature of our system and the users' primary focus on the end-effector motion, we employ model predictive control to regulate the end-effector trajectory. This approach enables whole-body coordination between the manipulator and the fully-actuated UAV, ensuring precise and efficient end-effector motion control. As discussed in previous sections, the complex interaction between the manipulator and the UAV is treated as the disturbance in the modeling stage, which introduces uncertainty in the nominal model used in the whole-body MPC. To mitigate the disturbances and model uncertainties, we integrate the L1 adaptive controller, ensuring robust disturbance compensation and accurate tracking performance. The diagram of the control algorithm is illustrated in Fig. \ref{fig: pipeline}.

\subsection{End-Effector-Centric Model Predictive Controller}

In the following, we describe the whole-body MPC framework used to optimize the end-effector reference trajectory. 

For the MPC formulation, we define the following state and control variables:
\begin{align}
\bm{x} := 
\begin{bmatrix}
\bm{p}_E &
{\bm{R}_{E}^W} &
\bm{v} &
\bm\theta
\end{bmatrix} \quad \bm{u}:= 
\begin{bmatrix}
\bm{\tau} &
\bm\theta_{\mathrm{cmd}}
\end{bmatrix} 
\end{align}

We use the following error functions for the position of the end-effector, the UAM velocity, the wrench control input, and the manipulator joint angle, respectively:
\begin{subequations}
\begin{align}
\bm{e}_p &= \bm{p}_E - \bm{p}_E^r\\
\bm{e}_R &= \frac{1}{2} \big( {\bm{R}_{W}^E}^r{}^\top \bm{R}_{W}^E - {\bm{R}_{W}^E}^\top {\bm{R}_{W}^E}^r \big)^\vee \\
\bm{e}_v &= \bm{v} - \bm{v}^r \\
\bm{e}_\theta &= \bm{\theta} - \bm{\theta}^r\\
\bm{e}_{u} &= \bm u - \bm u^r
\end{align}
\end{subequations}
where $(\bm{*})^\vee$ is the vee-operator that extracts a vector from a skew-symmetric matrix $\bm{*}$, and $(\cdot)^r$ represents the reference state values. The reference signals $\bm{p}_E^r$ and $\bm{R}_E^W$ are provided by a high-level teleoperation command or derived from an imitation learning policy. Manipulator default joint angle $\bm{\theta}^r$ is pre-selected, and reference control $\bm u^r = [\bm 0_3, \hat{\bm\theta}]$, where $\hat{\bm\theta}$ is the current joint states. 

The MPC formulation minimizes a cost function over a finite time horizon $H$ while subject to system dynamics and constraints:
\begin{subequations}
\begin{align}
\bm u_{\mathrm{opt}} = \arg\min_{\bm u} & \left\{ L_e(\bm x_H, \bm x^r_H) + \sum_{i=1}^{H}  L_r(\bm x_n, \bm x_n^r, \bm u_n)\right\} \label{cost}\\
\text{s.t.}\quad \bm x_{n+1} &= \bm f_{dyn}(\bm x_n, \bm \tau_n) \label{mpc_dyn} \\
\bm x_0 &=\hat{\bm x}, \quad
\bm x_n \in \mathcal{X} \label{mpc_state}\\ 
\bm u_{\text{lb}}&\leq \bm u \leq \bm u_{\text{ub}}\label{mpc_act}
\end{align}
\end{subequations}
Eq. \eqref{cost} defines the optimization objective, where $H$ represents the discrete prediction horizon. The stage and terminal costs, ${L}_r$ and ${L}_e$, are quadratic functions of the tracking errors, given by $\bm e_i^\top \bm Q_i \bm e_i$, where $\bm{x}_i \in \{\bm e_p, \bm e_R, \bm e_v, \bm e_\theta, \bm e_u\}$. The gain matrices $\bm{Q}_i$ are positive definite and tuned experimentally to balance precision and robustness.

Eq. \eqref{mpc_dyn} enforces the discrete-time system dynamics, incorporating the fully actuated UAV dynamics Eq. \eqref{uav_dyn}, manipulator kinematics Eq. \eqref{arm_kin}, and joint servo dynamics Eq. \eqref{arm_dyn}. The continuous system dynamics are discretized using a fourth-order Runge-Kutta (RK4) integration scheme to maintain numerical stability and accuracy. Disturbances $\bm \tau_{\mathrm{ext}}$ and $\bm d$ are ignored in the MPC formulation and solving process, but will be handled in the following Section \ref{L1} via online L1 adaptation.

Eq. \eqref{mpc_state} introduces state constraints, where $\hat{\bm{x}}$ represents the latest state estimate. The feasible state space $\mathcal{X}$ is defined by:
1) {Self-collision avoidance constraints}: ensuring that the manipulator does not collide with the UAV structure.  
2) {Environment collision constraints}: preventing the UAV contact with external obstacles.  
3) {Safety constraints}: including velocity limits and joint angle restrictions to ensure safe operation.

Eq. \eqref{mpc_act} imposes actuation limits on the aerial manipulator, where $\bm{u}_{\text{lb}}$ and $\bm{u}_{\text{ub}}$ define the lower and upper bounds of the control inputs.

The end-effector centric whole-body MPC formulation is a general framework that can adapt to various vehicle types and can extend to systems with multiple end-effectors. The control inputs, UAV dynamics, and arm kinematics can be tailored to specific vehicle and manipulator configurations, ensuring flexibility across different aerial manipulation systems.

\subsection{L1 Online Adaptation}\label{L1}

As discussed in previous sections, the complex interaction between the manipulator and the UAV is treated as the disturbance, which introduces uncertainty in the nominal model used in the whole-body MPC. Such model uncertainties are typically bounded, and prior knowledge about their characteristics is usually available. To mitigate the disturbances and model uncertainties, we integrate the L1 adaptive controller, ensuring robust disturbance compensation and accurate tracking performance.

We adopt the L1 adaptive controller from \citep{l1_drone} in both the fully-actuated UAV motion control and the manipulator joint angle tracking control, to compensate the disturbance $\bm\tau_{\mathrm{ext}}$ in Eq. \eqref{uav_dyn} and $\bm d$ in Eq. \eqref{arm_dyn}.

The adaptation law is designed by
\begin{subequations}
\begin{align}\label{uav_pred}
\bm{M} \dot{{\hat{\bm{ v}}}} &+ \bm{C} {\hat{\bm{ v}}} + \bm{g} = {\bm{\tau}} + {\hat{\bm{\tau}}}_{\mathrm{ext}} + \bm A_{v} (\hat{\bm{v}} - \bm{v})\\
\hat{\bm \tau}_{\mathrm{ext}}' &= -(e^{\bm{A}_v dt} - \bm I_{6\times 6})^{-1}\bm{A}_v e^{\bm{A}_v dt}(\hat {\bm v} - \bm v)\label{uav_d_diff}\\
\hat{\bm \tau}_{\mathrm{ext}} &\leftarrow \text{low pass filter}(\hat{\bm \tau}_{\mathrm{ext}}, \hat{\bm \tau}_{\mathrm{ext}}')\label{uav_d_lp}
\end{align}    
\end{subequations}
where $\hat{\bm v} \in \mathbf{R}^6$ denotes the estimated UAV velocity, $\bm A_v$ is a Hurwitz matrix, $dt$ is the discretization step length, and $\hat {\bm \tau}_{\mathrm{ext}} \in \mathbb{R}^6$ encapsulates the unknown wrench disturbances. Here Eq. \eqref{uav_pred} is a velocity estimator and Eq. \eqref{uav_d_diff} and Eq. \eqref{uav_d_lp} update and filter the disturbance $\hat{\bm \tau}_{\mathrm{ext}}$.  

Thus, the total UAV wrench control command $\bm \tau^*$ is computed as \begin{equation}
    \bm \tau^* = \bm \tau_{\mathrm{mpc}} + \hat{\bm \tau}_{\mathrm{ext}}
\end{equation} 

Similarly, the L1 adaptive controller for the manipulator joint angles is formulated to compensate for dynamic disturbances and model uncertainties:
\begin{subequations}
\begin{align}\label{arm_pred}
    &{\mathrm{diag}(\bm \beta)} \dot{{\hat {\bm\theta}}} + {\hat{\bm \theta}} = {\bm\theta}_{\mathrm{cmd}} + \hat{\bm d} + \bm A_d ({\hat{\bm \theta} - \bm \theta}),\\
    &\hat{\bm d}' = -(e^{\bm{A}_d dt} - \bm I_{4\times 4})^{-1} \bm{A}_d e^{\bm{A}_d dt}(\hat {\bm\theta} - \bm\theta), \label{arm_d_diff}\\
    & \hat{\bm d} \leftarrow \text{low pass filter}(\hat{\bm d}, \hat{\bm d}'). \label{arm_d_lp}
\end{align}    
\end{subequations}
where $\hat{\bm{\theta}}$ is the estimated joint state, and the disturbance term $\hat{\bm{d}} \in \mathbb{R}^4$. Thus, the final joint control command $\bm{\theta}^*$, incorporating the adaptive disturbance compensation, is given by:
\begin{equation}
    \bm{\theta}^* = \bm{\theta}_{\mathrm{cmd}} + \hat{\bm{d}}.
\end{equation}

\section{EE-Centric Teleoperation and Policy Learning}

\label{sec: drone_policy}

As we mentioned, our framework enables the decoupling between the high-level policy and low-level controller, while the interface between them is only the ee-centric interface. This allows the policy to be embodiment-agnostic, eliminating the need to consider low-level tracking control. In this section, we introduce two aerial manipulation system we developed based on this framework: ee-centric aerial teleoperation system and imitation-learning-based autonomous aerial manipulation system.

\subsection{EE-Centric Aerial Teleoperation}

We developed an aerial manipulation teleoperation system with the ee-centric interface, 
allowing the operator to focus solely on controlling the target end-effector pose, as if they had complete control of a freely moving hand in 3D space.

Robotic teleoperation requires bidirectional communication between the user and the robot. For the user-to-robot command, we developed a gamepad program so that the user can control the end-effector position ${\bm p_{E}}^r$ and orientation ${\bm R_{E}^{W}}^r$ with buttons and joysticks.

For robot-to-user communication, different from tabletop and mobile manipulation settings, in aerial manipulation, the user often lacks direct visual access to the workspace, necessitating reliance on onboard perception systems. In our work, we address this limitation by providing real-time visualization of RGB images captured from cameras mounted on both the end-effector and the base of the UAM. These images are displayed on monitors for continuous user observation. Further enhancing teleoperation efficacy, we have found it crucial to also visualize the user's inputs directly. To this end, we render the commanded target end-effector pose trajectories in real-time within 3D world frame plots, as illustrated in Fig. \ref{fig: pipeline}. This dual approach of visual feedback not only improves spatial awareness but also significantly enhances user performance in teleoperated tasks.

\subsection{EE-Centric Policy Learning}

To establish an autonomous aerial manipulating policy for versatile tasks, we develop an ee-centric policy learning framework based on imitation learning. Specifically, we adopt Action Chunk with Transformer (ACT) as the network structure \citep{zhao2023learning}. ACT utilizes a Conditional Variational Autoencoder (CVAE) where the encoder compresses action sequences and joint observations into a latent style variable. The transformer-based decoder generates action sequences from the latent variable (only during training and set to be the mean of the prior during testing), current joint observations, and encoded image features. The action chunking mitigates compounding errors and enhances the model's ability by predicting multiple future actions at once.

In this work, the ACT policy as well as policy observation and action are defined as follow: \begin{align}
    \bm a_{t:t+K} &= \pi_\varphi (\bm o_t)\\
    \bm o_t &= \left\{
        {I_E} , \ I_B , \ \bm p_{E} , \ {\bm R_{E}^{W}}_t
    \right\}\\
    \bm a_t &=\left\{{\bm p_{E}^r}, \ {{\bm R_{E}^{W}}^r}\right\}_t
\end{align} where $\pi$ denotes the ACT policy, $\varphi$ is the network parameter, $K$ is the chunking size, $I_B$, $I_E$ are RGB images from the base camera and the end-effector camera, each with $640 \times 480$ resolution. ${\bm p_{E}}$, ${\bm R_{E}^{W}}$, ${\bm p_{E}^r}$, ${{\bm R_{E}^{W}}^r}$ denotes the current and target UAM position and orientation, respectively. We use ResNet-18 as the backbone to encode the RGB images before inputting them into the transformer encoder. The flowchart of the algorithm implementation is illustrated in Fig. \ref{fig: pipeline}.

\section{Experiments}
\label{sec: experiment policy}

To validate the effectiveness of our proposed framework, we conduct a series of experiments focusing on end-effector trajectory tracking, aerial teleoperation, and policy learning for autonomous aerial manipulation\footnote{Please check out our project page for more visualization videos: \url{https://lecar-lab.github.io/flying_hand/}.}. Our goal is to assess whether the proposed whole-body MPC with L1 adaptation approach enhances trajectory tracking accuracy. Additionally, we evaluate how the ee-centric interface facilitates intuitive and precise teleoperation, reducing operator burden and improving task execution in a series of teleoperation tasks. Finally, we investigate whether the high-quality teleoperation demonstrations can be leveraged to train imitation learning-based policies for autonomous aerial manipulation in both simulation and real-world environments.

\subsection{Experimental Setup}

\subsubsection{Trajectory Tracking Task Setup}

To show the effectiveness of our proposed method in end-effector trajectory tracking tasks, we perform a comparison between our control methods against two baseline approaches:

\begin{itemize}
\item \textbf{w.o. MPC}: This baseline disregards both the fully-actuated UAV and arm dynamics, relying solely on inverse kinematics (IK) for motion planning. The UAV follows the desired trajectory generated by the IK planner using a cascade PID controller. 
\item \textbf{w.o. L1}: This baseline excludes the L1 adaptive component, leaving disturbances from UAV and manipulator interactions and modeling uncertainties uncompensated during control execution. 
\end{itemize}

We conduct experiments with three types of reference trajectories for the end-effector, each lasting 60 seconds. The setpoint trajectory requires the aerial manipulator to keep the end-effector hover at a fixed position $\bm p_E = [0.0, 0.0, 1.3]$. The ellipse trajectory requires to track a sinusoidal trajectory defined as $\bm p_E = [0.5\sin(0.3t), 0.0, 1.4+0.2\sin(0.3t+0.75)]$. The figure-8 trajectory requires to track a trajectory $\bm p_E = [0.1 + 0.6\sin(0.3t), 0.0, 1.35+0.25\sin(0.6t)]$. The maximum velocity in the reference trajectory is about $0.2$ m/s. The pitch, row and yaw attitude of the end-effector is fixed at zero during the tracking. Root Mean Square Error (RMSE) is used as the tracking performance evaluation criterion. Each trajectory is repeated three times to compute the mean and standard deviation.

\subsubsection{Aerial Manipulation Task Setup}

We conducted a series of experiments to evaluate the capabilities and applications of our aerial manipulation system. We select different typical tasks from each category we discussed in Sec. \ref{sec: related_works_am}, including:

\begin{itemize}

\item \textbf{Aerial Writing}: Drawing a target shape (the digit ``2025'') on a vertical wall, with an overall size of approximately $3 \text{m} \times 0.8 \text{m}$. This task required precise specification and tracking of the end-effector (EE) pose trajectory while maintaining stable contact with the surface.
\item \textbf{Aerial Peg-in-Hole}: Inserting a 20mm diameter pole into a 50mm diameter hole positioned around 150cm above the ground.
\item \textbf{Rotate Valve}: Manipulating the articulated valve by grasping its handle and rotating it along a 20cm diameter circle, emulating industrial valve manipulation.
\item \textbf{Aerial Pick and Place}: Grasping and placing various objects with different shapes and sizes including the screwdriver, pen, tape, and glue bottle.
\item \textbf{Unmount Light Bulb}: Grasping a mounted light bulb and unscrewing it out of the socket.
\item \textbf{Mount Light Bulb}: A long horizon task that requires a sequence of motion, including inserting a light bulb into a socket, screwing it in, and subsequently turning it on by pressing the button.
\end{itemize}

For different tasks, different end-effectors are adopted, including the parallel jaw gripper for the pick and place and peg-in-hole task, a passive elastic claw for grasping the light bulb, and a bucket-shaped gripper for rotating the valve.

\subsection{Implementation Details}

The optimal control problem in the ee-centric MPC is implemented using ACADOS~\citep{Verschueren2021} with a $25$ms discretisation step and a $2.5$s constant prediction horizon, running in $100$ Hz, and other controller parameters are listed in Table \ref{tab:mpc_param}. The control output is executed in a receding horizon style, where at each iteration, only the first control input $\bm u_0$ is applied to the system.

\begin{table}[t]
\centering
\caption{\textbf{Controller Parameters}}
\label{tab:mpc_param}
\renewcommand{\arraystretch}{1.2}
\begin{tabular}{l p{5.5cm}}
\toprule
Horizon Length $T$ & $2.5$ s \\
Horizon Steps $N$ & $100$ \\
State Cost $\bm{Q}_p$ & $\mathrm{diag}(12, 12, 12)$ \\
Rotation Cost $\bm{Q}_R$ & $\mathrm{diag}(10, 10, 10)$ \\
Velocity Cost $\bm{Q}_v$ & $\mathrm{diag}(0.1, 0.1, 0.1)$ \\
Joint Angle Cost $\bm{Q}_\theta$ & $\mathrm{diag}(0.1, 0.1, 0.1)$ \\
Control Cost $\bm{Q}_u$ & $\mathrm{diag}(0.03, 0.03, 0.03, 0.1, 0.1, 0.1)$ \\
\bottomrule
\end{tabular}
\end{table}

Since both the L1 adaptive controller and the MPC controller require accurate system modeling $\bm f_{dyn}$ to achieve effective control performance, we perform system identification to estimate uncertain parameters shown in Table \ref{tab:sysid}.
We excite the system with two types of motions. First, arm motion-only trajectories are executed while keeping the UAV stationary to calibrate the DH parameters $\bm \zeta$ and joint servo delay $\bm\beta$. These trajectories ensure that the manipulator kinematic parameters and joint motor dynamics accurately reflect the actual manipulator motion response. Second, UAV free-flight trajectories are conducted to identify the drone dynamics described in Eq. \eqref{uav_dyn}.

\begin{table}[t]
\centering
\caption{\textbf{System Identification Results}}
\label{tab:sysid}
\renewcommand{\arraystretch}{1.2}
\begin{tabular}{l p{5.5cm}}
\toprule
Mass Matrix $\bm{M}$ & $\mathrm{diag}(0.105, 0.121, 0.101, 0.025, 0.011, 0.013)$ \\
Motor Delay $\beta$ & ($0.66$, $0.68$, $0.81$, $0.85$) \\
Joint 1 DH Param $\bm \zeta_1$& $d_1 = 0.0$, $a_1 = 0.363$, $\alpha_1 = 0.10$ \\
Joint 2 DH Param $\bm \zeta_2$ & $d_2 = 0.050$, $a_2 = 0.441$, $\alpha_2 = -0.10$ \\
Joint 3 DH Param $\bm \zeta_3$ & $d_3 = 0.0$, $a_3 = 0.007$, $\alpha_3 = -1.578$ \\
Joint 4 DH Param $\bm \zeta_4$ & $d_4 = 0.076$, $a_4 = 0.200$, $\alpha_4 = 0.0$ \\
\bottomrule
\end{tabular}
\end{table}

\subsection{Experiment I: Trajectory Tracking}

Table \ref{tab:tracking} shows the comparison of our approach against w.o. MPC and w.o. L1 baselines in three types of reference end-effector trajectories. The results show that our proposed method achieves the lowest tracking error, with approximately 1 cm in hover and 4 cm during motion. In contrast, the baseline w.o. L1  exhibits 1.3 cm and 6.5 cm, respectively, while the baseline w.o. MPC performs the worst, with 2 cm in hover and 10 cm in motion. The increased tracking error is primarily due to the complex interaction between the UAV and the manipulator, where simultaneous movement introduces dynamic coupling effects that make precise tracking more challenging.

\begin{figure}
    \centering
    \includegraphics[width=1.0\linewidth]{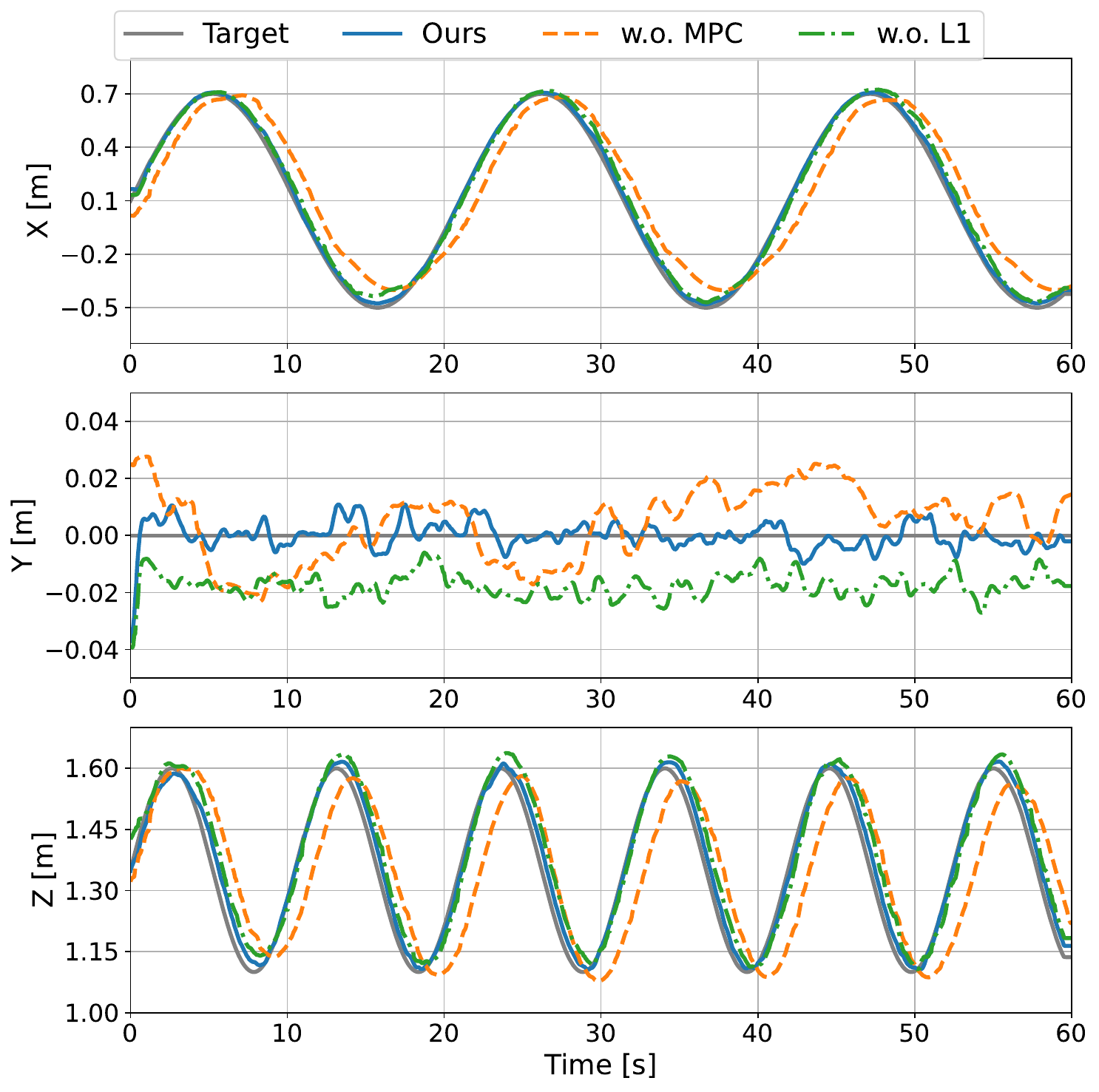}
    \caption{End-effector tracking performance of aerial manipulator in Ellipse trajectory. Tracking results indicate that the w.o. MPC baseline exhibits significant tracking lag, while the w.o. L1 baseline suffers from static tracking errors due to model mismatches. }
    \label{fig: tracking}
\end{figure}

\begin{figure}
    \centering
    \includegraphics[width=1.0\linewidth]{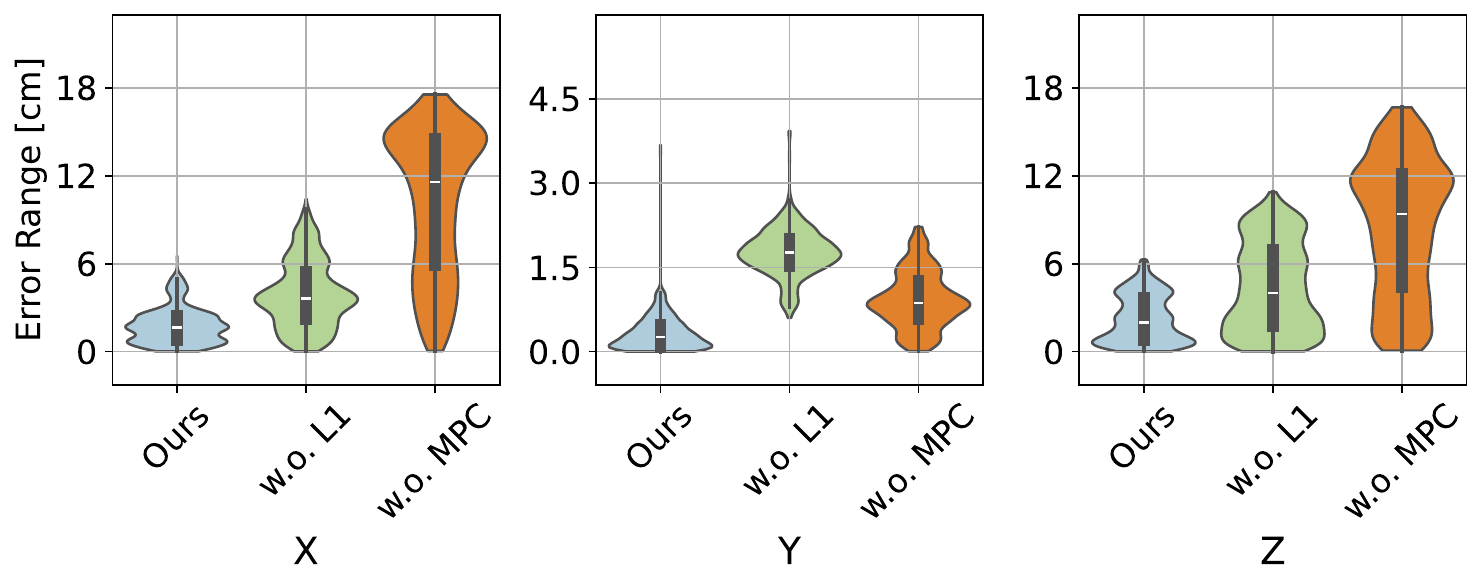}
    \caption{End-effector tracking error distribution for three types of trajectories using our methods and two baselines.
      }
    \label{fig: err_dist}
\end{figure}

Fig. \ref{fig: tracking} shows that our method (blue) demonstrates the best tracking performance. The baseline w.o. L1  (green) exhibits overshoot in the X and Z axes and bias in Y, indicating that the L1 controller effectively mitigates both transient and steady-state errors caused by model uncertainties. The baseline w.o. MPC  (orange) suffers from significant motion lag, as its inverse kinematics fail to account for UAV dynamics. Fig. \ref{fig: err_dist} depicts the error distribution for all trajectories using the proposed methods and two baselines, showing that our method achieves the smallest and most centered error, whereas the L1 baseline exhibits steady-state errors and the MPC baseline displays a broader error spread due to dynamic lag. These results confirm the effectiveness of our proposed control scheme. 

\begin{figure}[h]
    \centering
    \includegraphics[width=1.0\linewidth]{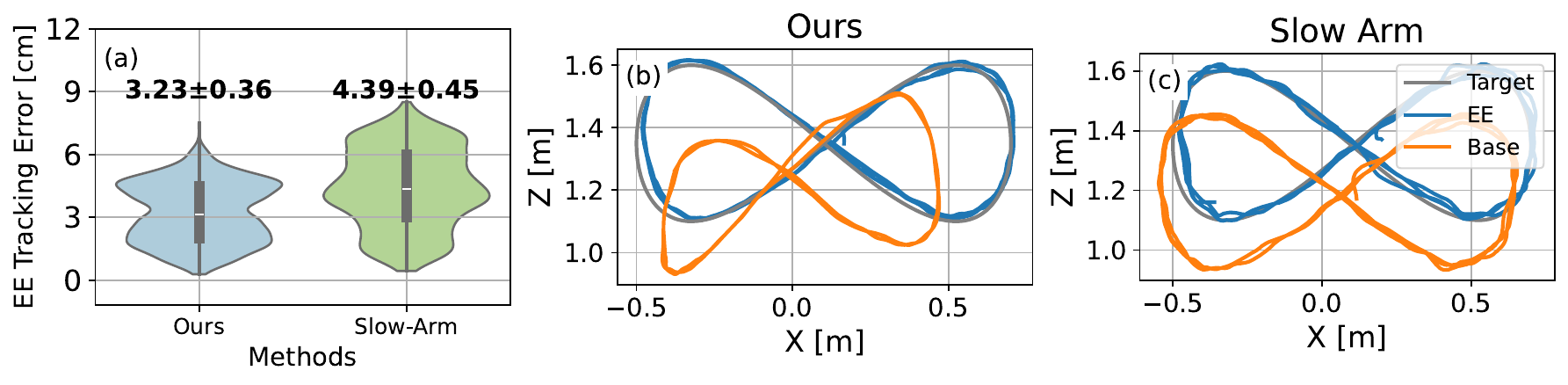}
    \caption{Arm flexibility ablation study for MPC controller}
    \label{fig: err_dist_arm}
\end{figure}

We also investigate the contribution of arm flexibility to EE tracking performance by increasing the arm control cost $5$ times larger. Fig.~\ref{fig: err_dist_arm}a \emph{Slow-Arm} results show a 35\% larger EE tracking error when arm flexibility is restricted, with a more oscillating EE trajectory (Fig.~\ref{fig: err_dist_arm}c) compared to the MPC with flexible arm (Fig.~\ref{fig: err_dist_arm}b). This aligns with the intuition that higher arm flexibility improves EE tracking performance, as the arm typically responds faster than the drone base. 

\begin{figure}[h]
    \centering
    \includegraphics[width=1.01\linewidth]{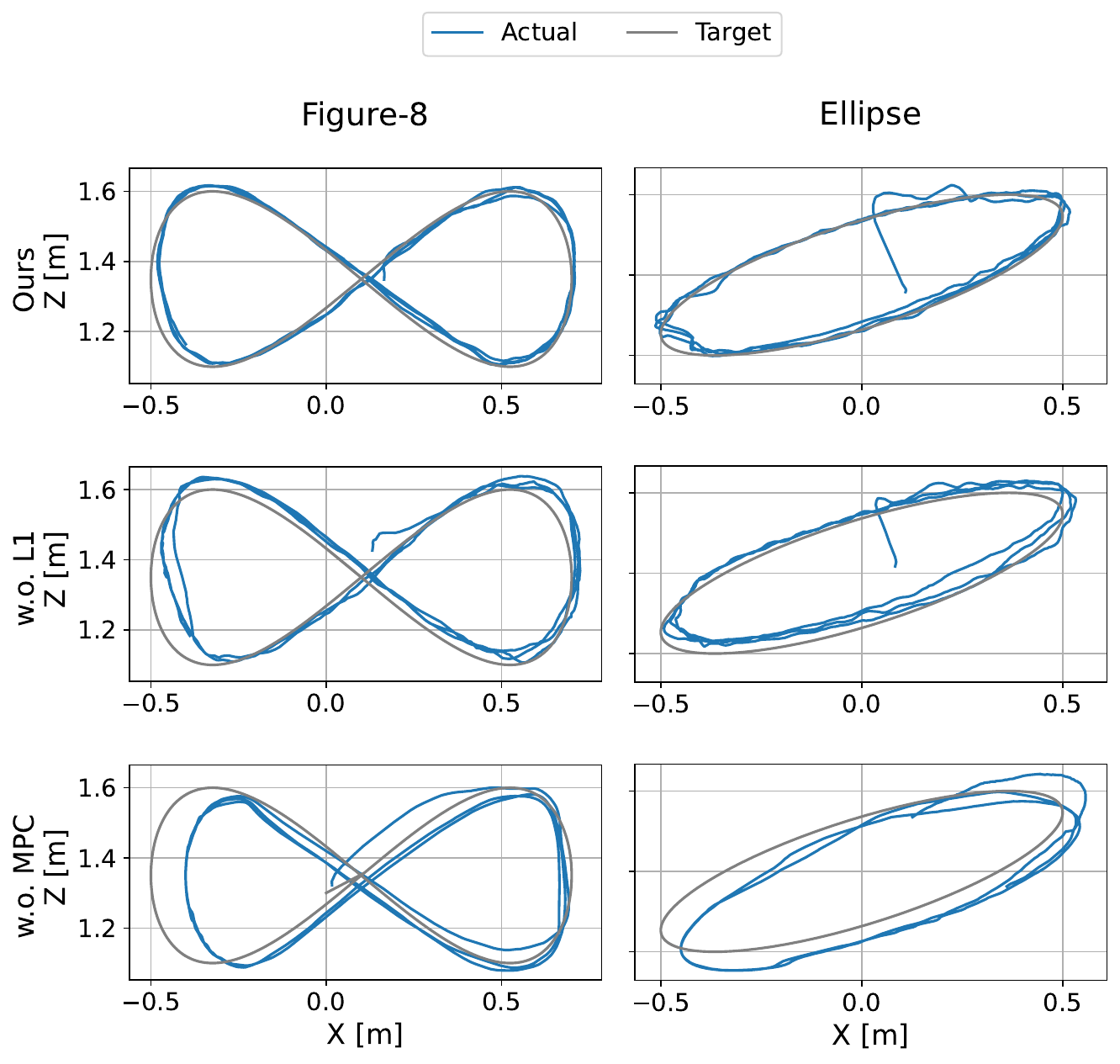}
    \caption{Comparison of Figure-8 and Ellipse trajectory tracking performance across three methods. Our approach achieves the lowest tracking error in dynamic trajectory tracking tasks.}
    \label{fig: x-z}
\end{figure}

Despite the high tracking performance of the proposed methods, Fig. \ref{fig: x-z} reveals that tracking error increases at lower altitudes (around $1$m), likely due to unmodeled ground and wall effect disturbances. Additionally, oscillations in the end-effector trajectory are observed across all methods. We notice servo backlash (around $0.5^\circ$ dead zone), which results in a 2 cm control dead zone in the end-effector task space, limiting tracking precision during fast UAV maneuvers. Further improvements can be achieved through more accurate system modeling and higher-precision hardware to enhance tracking accuracy.

\begin{table}[t]
\centering
\caption{End-Effector Trajectory tracking performance}
\label{tab:tracking}
\begin{tabular}{lcccc}
\toprule
RMSE (cm) & Setpoint & Ellipse & Figure-8 &  \\
\midrule
Our Method & \textbf{1.00 $\pm$ 0.11} & \textbf{3.98 $\pm$ 0.41} & \textbf{4.62 $\pm$ 0.58}\\
w.o. L1 Adaptation & 1.33 $\pm$ 0.16 & 6.67 $\pm$ 0.42 & 6.28 $\pm$ 0.50 \\
w.o. MPC &  2.07 $\pm$ 0.19 & 11.25 $\pm$ 0.83 & 9.64 $\pm$ 1.07 \\
\bottomrule
\end{tabular}
\end{table}

\subsection{Experiment II: Aerial Teleoperation}

\begin{figure}
    \centering
    \includegraphics[width=0.99\linewidth]{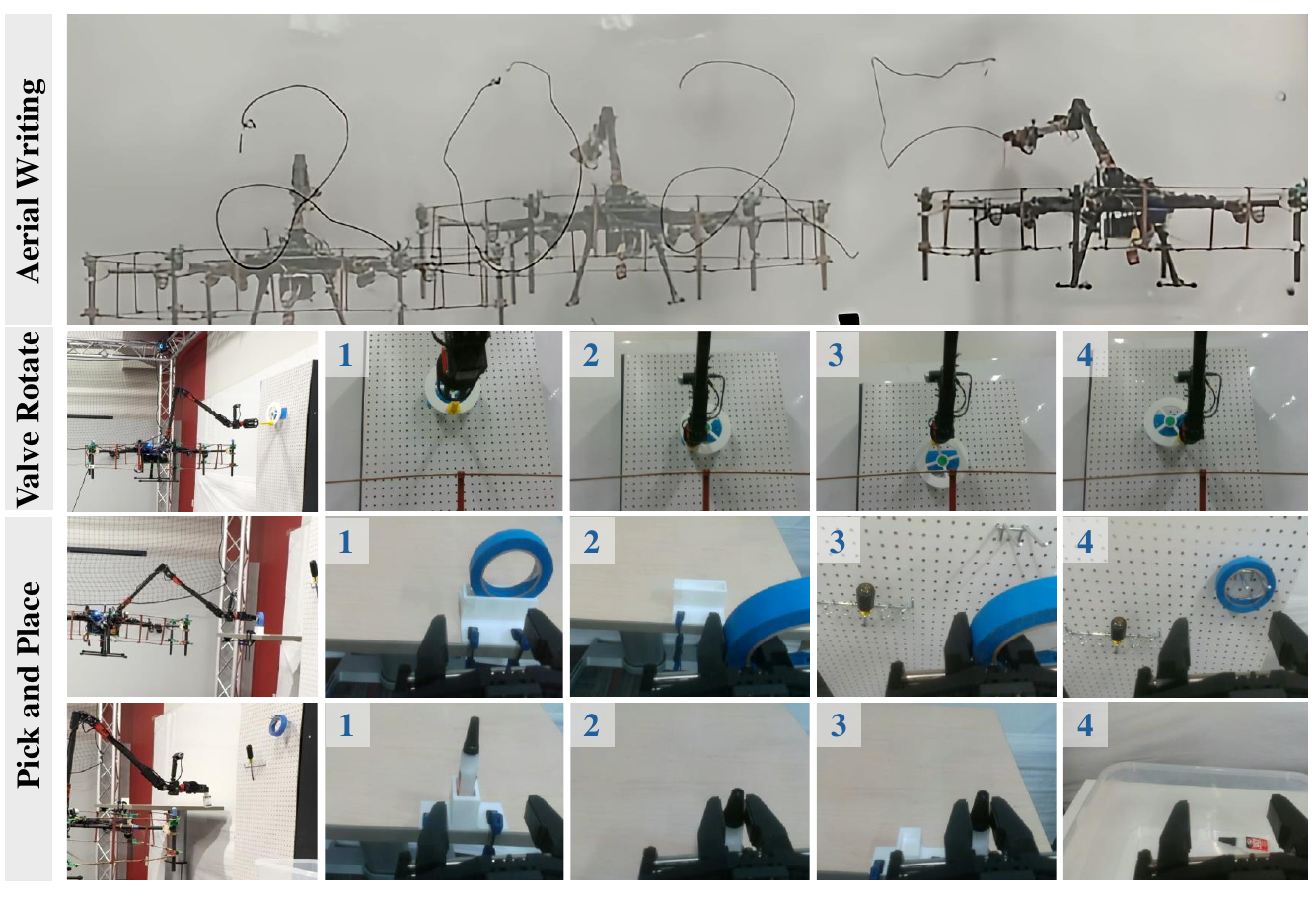}
    \caption{Aerial Teleoperation Manipulation Tasks. We target 1) Aerial Writing: UAM with a marker pen writes '2025' on a whiteboard. 2) Rotate Valve: UAM grasps the handle and rotates the valve with one loop. 3) Pick-and-Place: UAM grasps an object and relocates it to a designated area.}
    \label{fig: tasksetup}
\end{figure}

\begin{figure*}
    \centering
    \includegraphics[width=0.99 \linewidth]{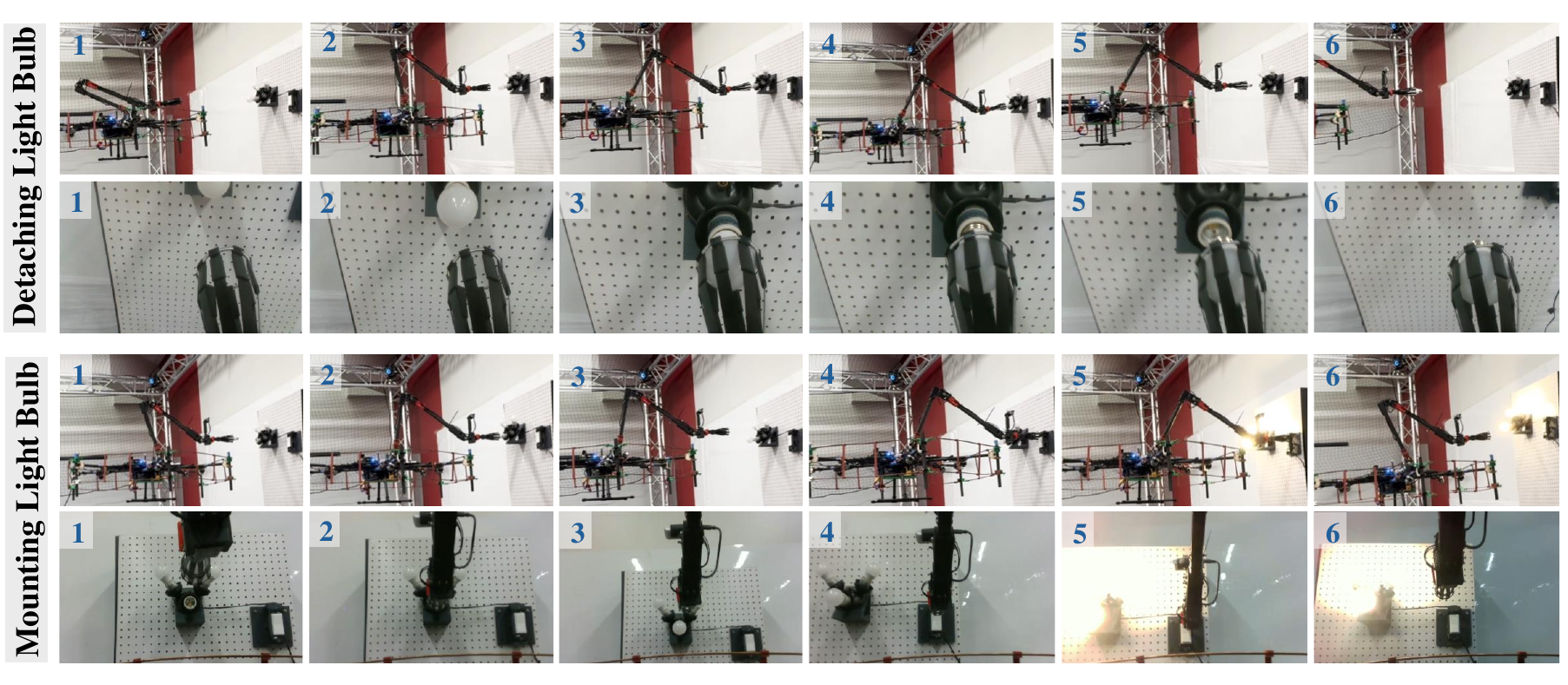}
    \caption{Long horizon aerial teleoperation light bulb changing task. UAM grasps the light bulb and unscrews it down during the first flight. And it inserts, screws a new light bulb, and presses the button to turn on the light during the second flight.}
    \label{fig: lightbulb_change}
\end{figure*}

First, we demonstrated the effectiveness and versatility of the proposed teleoperation system, by targeting aerial writing, rotating the valve, aerial pick and place, unmount, and mount light bulb tasks. As shown in Fig. \ref{fig: tasksetup} and Fig. \ref{fig: lightbulb_change}, human teleoperators can easily achieve all aerial manipulation tasks with little learning and operation cost. One key success factor we attribute is the ee-centric interface, which reduces human effort and improves the quality of teleoperation data for future policy learning.

\begin{figure}
    \centering
    \includegraphics[width=0.99\linewidth]{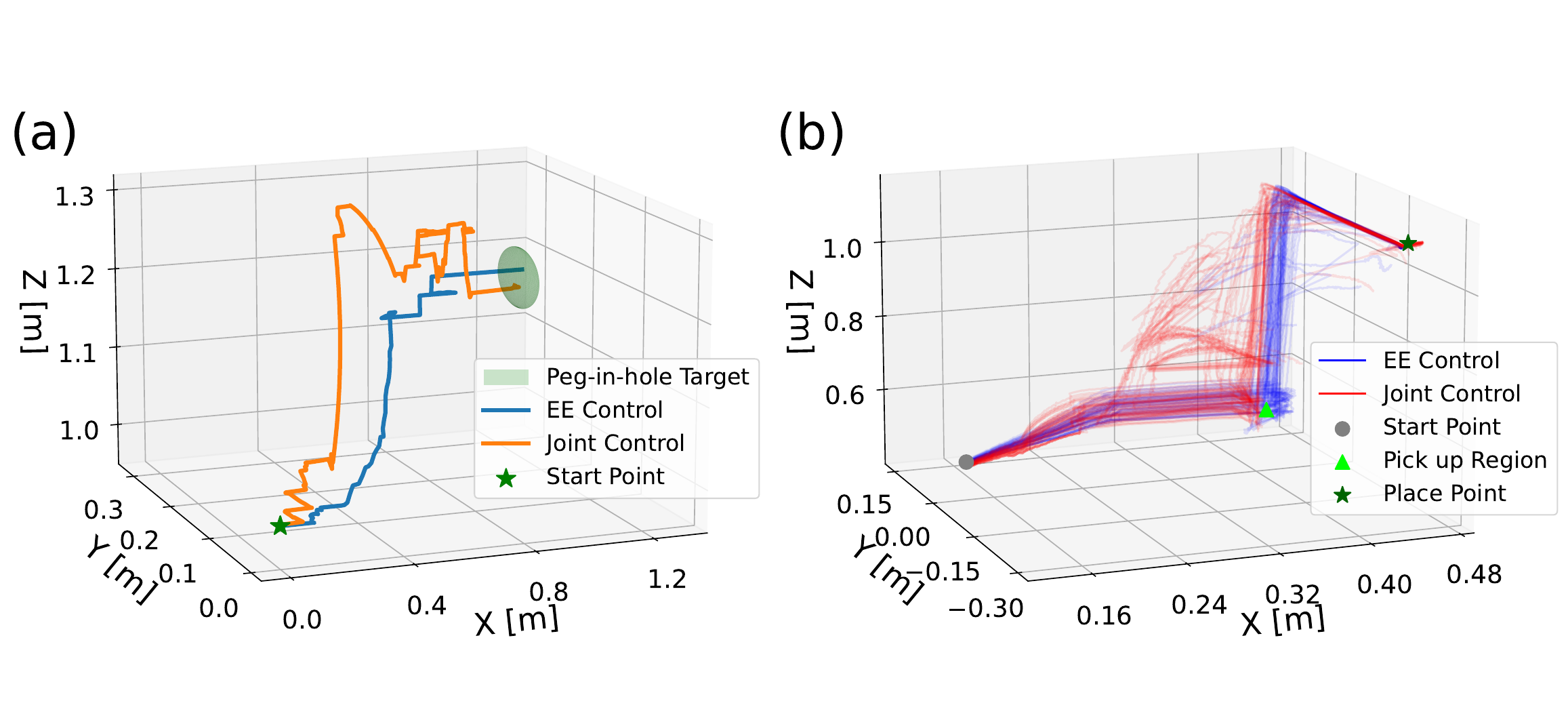}
    \caption{(a). End-effector command trajectory using the ee-centric teleoperation interface and full-DoF teleoperation interface, in a simulated peg-in-hole task. (b). End-effector command trajectory of the learned autonomous policy during 50 test trials using ee-centric interface and using full-DoF interface, in a simulated pick and place task.}
    \label{fig: eevsjoint}
\end{figure}

In a subsequent experiment, we evaluated the benefits of directly controlling the end-effector's pose using our framework against controlling each degree of freedom (DoF) for UAVs and robotic arms \citep{yashin2019aerovr} in a simulated peg-in-hole task. The teleoperation command trajectories are illustrated in Fig. \ref{fig: eevsjoint}a.  Direct control of the end-effector allowed operators to issue more fluid command trajectories, significantly enhancing the precision of end-effector movements and decreasing the time required to complete the task.

\subsection{Experiment III: Learning from Demonstration}

\subsubsection{Simulation Experiments}

\begin{figure}[t]
    \centering
    \includegraphics[width=0.9\linewidth]{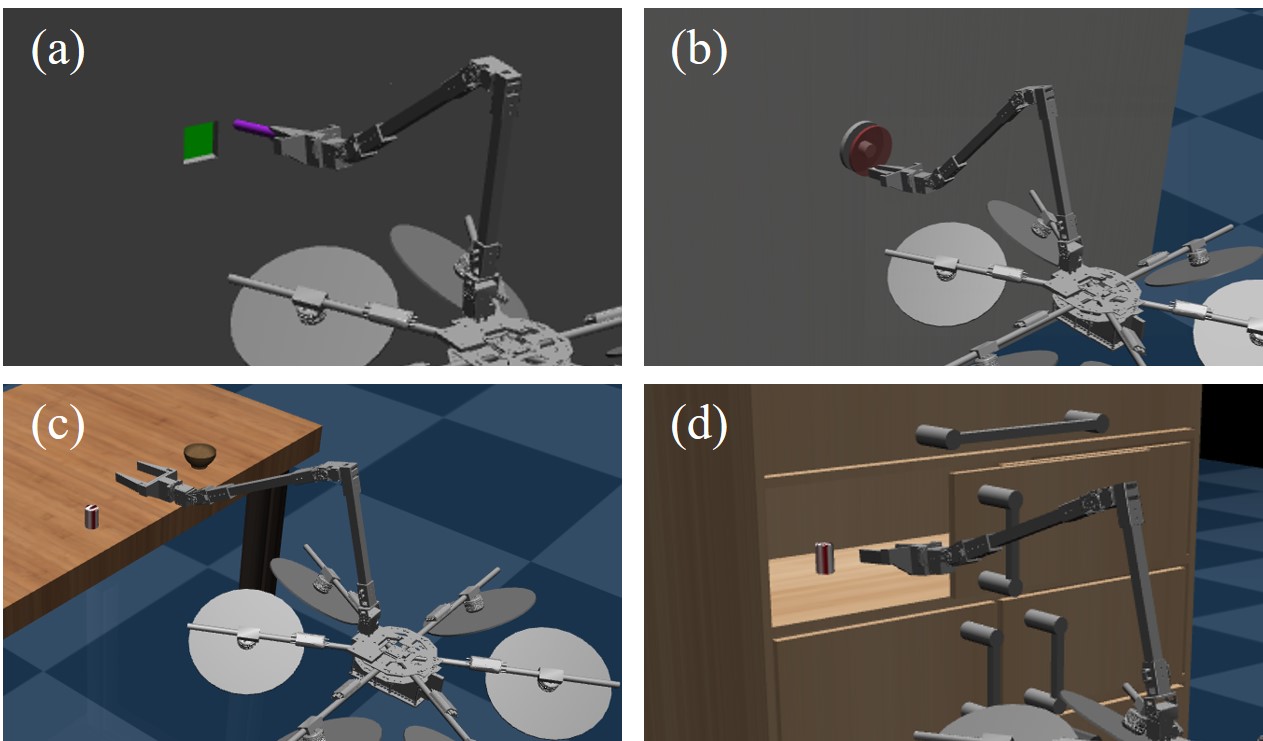}
    \caption{Task setup in Mujoco simulation, including (a) Peg-in-Hole; (b) Rotate the Valve; (c) Pick and Place; and (d), a long horizon Open and Retrieve task.}
    \label{fig:mujoco_sim}
\end{figure}

We first demonstrate our learning from demonstration framework in Mujoco~\citep{mujoco2012} simulator with four tasks: (i)~\textit{peg-in-hole}, (ii)~\textit{rotate valve}, (iii)~\textit{pick and place} and (iv)~\textit{open and retrieve}, as shown in Fig.~\ref{fig:mujoco_sim}~(a). To collect demonstrations, we use a scripted policy for each task. Every episode of the scripted policy lasts about 12 seconds for each task. We collect 50 episodes for each task and the control frequency is 50 Hz. 
Note that with our ee-centric interface, we do not consider any joint configuration when collecting demonstrations, which allows us to efficiently collect smooth demonstrations without tediously adjusting each joint position to complete the task. Our ACT policy for each task in simulation is trained with 100 action chunk size and limited 5k epochs. After training, we choose the policy with the least validation loss to perform 50 evaluation trials. The evaluation result is shown in Table~\ref{tab:act_sim}.

To showcase the advantage of learning from an ee-centric demonstration compared to a joint space demonstration, we use the same demonstration trajectory but change the observation and action to be in UAM configuration space, i.e., the UAV position and orientation, and each joint angle of the robotic arm. After that, we train a joint space ACT policy with the same training setting as the ee-centric ACT policy, except that the EE pose in the observation and action space is replaced by the drone base pose and full manipulator joint angles. We compare their success rate in evaluation summarized in Table~\ref{tab:act_sim}. It shows that with limited 5k training epochs, ee-centric policy outperforms the joint space policy in challenging tasks including \textit{pick and place}, \textit{peg in hole}, and \textit{open and retrieve}. As illustrated in Fig. \ref{fig: eevsjoint}b, the ee-centric policy targets the pickup region and the place point more precisely, while the joint space policy is more prone to generate wrong targets.

Overall, the experimental results reveal several complementary insights:

\begin{itemize}
    \item \textbf{Geometric Precision Advantage}: Our ee-centric policy achieves 2.5$\times$ higher success rate in geometrically sensitive \textit{peg in hole} task, directly benefiting from task-space supervision that eliminates the accumulated EE error from the joint space.
    \item \textbf{Multi-Skill Composition}: In the \textit{open and retrieve} task, our ee-centric policy achieves  2$\times$ higher success rate than joint space policy, which demonstrates its inherent advantages in multi-skill decomposition and execution.
\end{itemize}

\begin{table}[t]
\centering
\caption{Imitation Learning Simulation Success Rate}
\label{tab:act_sim}
\setlength{\tabcolsep}{1.pt}
\small
\begin{tabular}{@{}l|c|c|c|c@{}}
\toprule
 & \rotatebox{0}{Rotate Valve} 
 & \rotatebox{0}{Pick \& Place} 
 & \rotatebox{0}{Peg in Hole} 
 & \rotatebox{0}{Open \& Retrieve} \\
\midrule
Joint Space & 50/50 & 38/50 & 9/50 & 8/50 \\
EE  & 50/50 & \textbf{48/50} & \textbf{23/50} & \textbf{17/50} \\

\bottomrule
\end{tabular}
\end{table}

\subsubsection{Real-world Experiments}
\begin{figure*}
    \centering
    \includegraphics[width=0.99\linewidth]{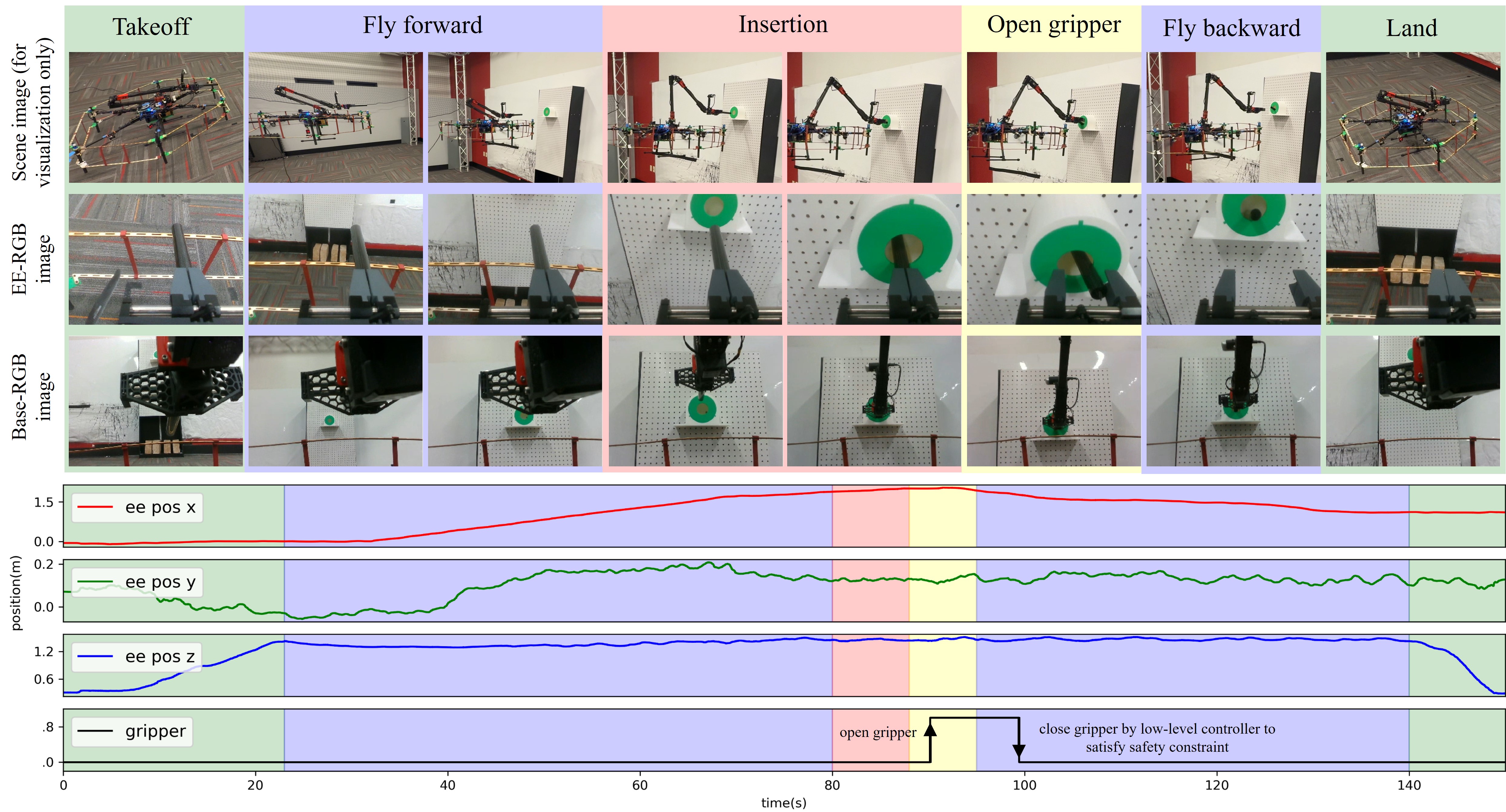}
    \caption{Autonomous aerial manipulation peg-in-hole policy experiment. The UAM inserts the peg precisely, highlighting both the accuracy of the learned policy and the low-level controller. }
    \label{Fig: act_peg_in_hole}
\end{figure*}

\begin{figure}
    \centering
    \includegraphics[width=1.0\linewidth]{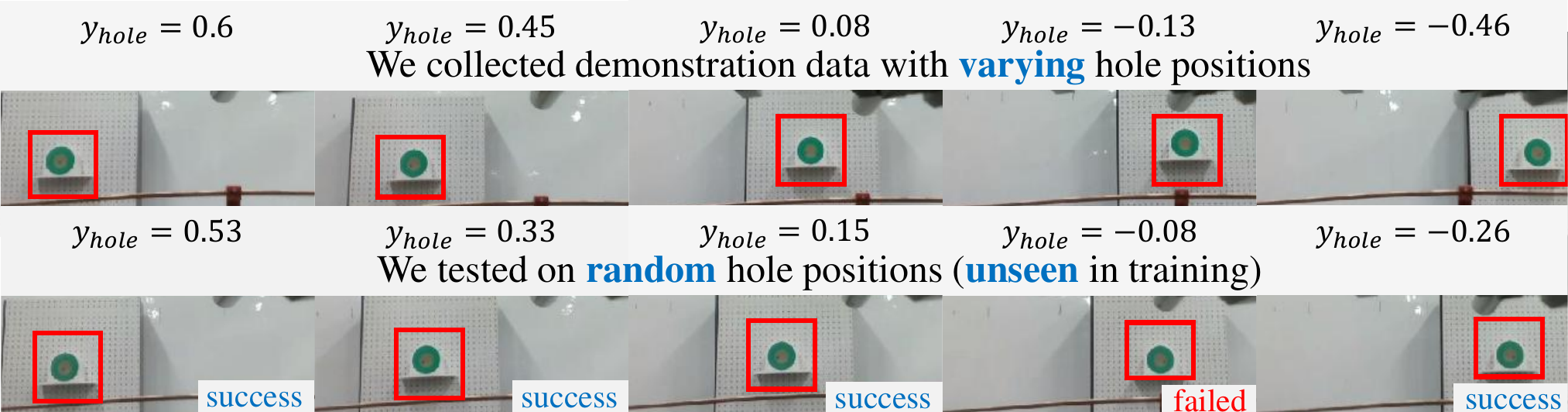}
    \caption{Task scenario randomization for peg-in-hole data collection and test.}
    \label{fig: experiment_real_peg_in_hole_scenario}
\end{figure}

We adopt the aerial peg-in-hole task to demonstrate our capability to derive an autonomous policy from human demonstrations for aerial manipulation in the real world. The task configurations are illustrated in Fig. \ref{Fig: act_peg_in_hole} and Fig. \ref{fig: experiment_real_peg_in_hole_scenario}. 

We collected 25 episodes of demonstration data via human teleoperation with varying hole horizontal positions, with each episode taking approximately 2 minutes, culminating in a total of around 50 minutes of operational data and about 2 hours of wall-clock time. The data was downsampled to 10 Hz, and the action chunk size was empirically set to 100 during the training process. After training through 100,000 epochs, the policy with the least validation loss is selected. We tested with random unseen horizontal hole positions and the learned policy successfully completed 4 out of 5 real-world peg-in-hole tests, i.e., 80\% successful rate. The UAM pushed the peg forward to the edge of the hole and didn't insert it inside successfully. These results underline the potential of learning-based approaches in aerial manipulation under our developed framework, though also highlight the need to develop robust recovery policies, especially for aerial manipulation.

\subsection{Discussion}

The experiments demonstrate our framework in end-effector trajectory tracking, aerial teleoperation, and policy learning for autonomous aerial manipulation. The precise end-effector control framework demonstrated superior end-effector tracking accuracy with minimal error. The high-precision control enables efficient user-friendly aerial teleoperation, allowing human operators to perform multiple complex tasks, which also helps high-quality demonstration data collection. Leveraging the ee-centric framework, advanced high-level policies such as imitation learning can be easily incorporated into the aerial manipulation field, which further expands the field development.

\section{Limitations}
\label{sec: limitations}

Although we have demonstrated the proposed framework through various real-world experiments, there are still several limitations due to time constraints and methodological limitations. Firstly, all of our experiments were conducted indoors within a motion capture system, where we achieved millimeter-level state estimation of the UAV and end-effector states using forward kinematics. This setup limits its practical application in real-life scenarios. Secondly, the current safety constraints are model-based, which further restricts practical applications. Incorporating onboard perception to detect obstacles and generate safety constraints in real-time will be our next step, as various studies have demonstrated the feasibility of UAV collision-free flight. Thirdly, the current performance is limited by the robotic arm actuators, which have relatively large backlash and would generate unavoidable vibrations. Finally, although the proposed framework, which decouples different modules, demonstrates the potential for cross-platform compatibility and integration with general manipulation fields, more real-world experiments are planed to take to further validate its effectiveness.

\section{Conclusion}
\label{sec: conclusion}

This work presents a unified aerial manipulation framework with the ee-centric interface for versatile aerial manipulation tasks. Our system includes a versatile hardware platform that consists of a fully actuated UAV and a 4-DOF manipulator, an ee-centric whole-body MPC to ensure precise end-effector tracking, and an ee-centric high-level policy, where we developed both an intuitive teleoperation and an imitation learning-based autonomous policy. Through extensive real-world experiments, we demonstrated the system's versatility across various aerial manipulation tasks, including writing, peg-in-hole, pick-and-place, valve rotating, and light bulb replacement. More importantly, we demonstrate how this modular and standardized EE-centric framework effectively decouples the high-level policy from the low-level controller, which enables seamless integration of existing standard high-level policy modules from the broader manipulation community, such as teleoperation and imitation learning, into the field of aerial manipulation. The proposed framework achieved high precision, adaptability, and robust performance, making it a significant step toward standardizing aerial manipulation within the broader manipulation field. Future work will extend the framework's applicability to outdoor environments, incorporate onboard perception for obstacle avoidance, and further improve the end-effector tracking performance.


\bibliographystyle{IEEEtran}
\bibliography{root}

\end{document}